\documentclass[journal]{IEEEtran}

\usepackage{times}
\usepackage{epsfig}
\usepackage{graphicx}
\usepackage{subfigure}
\usepackage{amsmath}
\usepackage{amssymb}
\usepackage{amstext}
\usepackage{url}
\usepackage{multirow}
\usepackage{algorithm}
\usepackage{algorithmicx}
\usepackage{algpseudocode}
\usepackage{xcolor}
\usepackage{epstopdf}
\usepackage{textcomp}
\usepackage{float}
\usepackage{lineno}
\usepackage{pifont}

\usepackage{multirow}
\usepackage{booktabs}
\usepackage{balance}
\usepackage{CJK}
\usepackage{color}
\usepackage{acronym}
\usepackage{xspace}

\usepackage[colorlinks,linkcolor=black, citecolor=black]{hyperref}

\begin{document}

\title{Hybrid Routing Transformer for Zero-Shot Learning}

\author{De~Cheng\dag,
    Gerong~Wang\dag,
    Bo Wang,
    Qiang Zhang\ding{41},
    Jungong Han,
    Dingwen Zhang.

\thanks{De~Cheng, Gerong Wang, Qiang Zhang, Xidian University, Xi'an, Shaanxi, P.R. China, Bo Wang is with the Tsinghua University, Beijing, P.R. China, Jungong Han is with the Aberytwyth University, Aberystwyth, UK, Dingwen Zhang is with the Northwestern Polytechnical University, Xi'an, Shaanxi, P.R. China.}
\thanks{\dag De Cheng and Gerong Wang are co-first authors.}
}
%



\maketitle

\begin{abstract}
 Zero-shot learning (ZSL) aims to learn models that can recognize unseen image semantics based on the training of data with seen semantics. Recent studies either leverage the global image features or mine discriminative local patch features to associate the extracted visual features to the semantic attributes. However, due to the lack of the necessary top-down guidance and semantic alignment for ensuring the model attending to the real attribute-correlation regions, these methods still encounter a significant semantic gap between the visual modality and the attribute modality, which makes their prediction on unseen semantics unreliable. To solve this problem, this paper establishes a novel transformer encoder-decoder model, called hybrid routing transformer (HRT). In HRT encoder, we embed an active attention, which is constructed by both the bottom-up and the top-down dynamic routing pathways to generate the attribute-aligned visual feature. While in HRT decoder, we use static routing to calculate the correlation among the attribute-aligned visual features, the corresponding attribute semantics, and the class attribute vectors to generate the final class label predictions. This design makes the presented transformer model a hybrid of 1) top-down and bottom-up attention pathways and 2) dynamic and static routing pathways. Comprehensive experiments on three widely-used benchmark datasets, namely CUB, SUN, and AWA2, are conducted. The obtained experimental results demonstrate the effectiveness of the proposed method.

\end{abstract}


\section{Introduction}

Deep learning has made great progress in a variety of vision tasks when the models are trained on large-scale labeled datasets. However, the real-world natural images follow a long-tailed distribution so that the data-hungry characteristic of CNN-based models limits their ability to recognize rare object classes, specially for the fine-grained animal species~\cite{han2020learning}. Meanwhile, an increasing number of newly defined visual concepts and products come to the fore so quickly, and the speed of data annotation for model training cannot keep up with the pace of new things emerging, thus the CNN-based models cannot be generalized to these new classes for testing. In contrast, a child can learn from only a few samples, summarize knowledge, and even draw inferences about other cases from one instance to recognize unseen objects. Therefore, building zero-shot learning (ZSL) models to transfer knowledge from seen classes to unseen classes is  significant and indispensable.

Zero-shot learning (ZSL) mimics the human ability to recognize objects only from a description in terms of concepts in some semantic vocabulary~\cite{morgado2017semantically}, and aims to recognize the unseen classes, of which the labeled images are unavailable during model training~\cite{lampert2013attribute,xian2017zero}. Existing works on ZSL mainly leverage the global features~\cite{akata2015evaluation} or patch features~\cite{xie2020region,xu2020attribute} to construct visual-semantic alignment models. In these approaches, images and attributes are embedded with compatibility function. Despite good performances on coarse-grained datasets (e.g., Animal with Attribute dataset~\cite{lampert2013attribute}), these approaches gradually degenerate when dealing with fine-grained datasets (e.g., Caltech-UCSD Birds-200-2011 dataset~\cite{welinder2010caltech}), since much more local discriminative information is required to distinguish these fine-grained classes. Several recent works~\cite{zhu2019semantic,xie2019attentive,ji2018stacked,ZhangScrible2021,ZhangSurvey2021} try to focus on discriminative visual feature learning, by introducing attention mechanism into zero-shot classification problem, such as the spatial and channel attention~\cite{zhu2019semantic}, region attention~\cite{xie2019attentive,ZhangWeakly2020}.
However, there still exists significant semantic gap between the visual modality and the attribute modality in the existing passive attention mechanisms as these methods generate attention weights purely in the bottom-up forward passing manner, which lacks the necessary top-down guidance and semantic alignment for attending to the real attribute-correlation regions. In this paper, We reveal a fundamental issue that establishing an active connection between the visual feature and attribute vector rather than a simple passive link is a key point to facilitate zero-shot learning.

\begin{figure*}[t]
	\centering
	\includegraphics[width=0.95\textwidth]{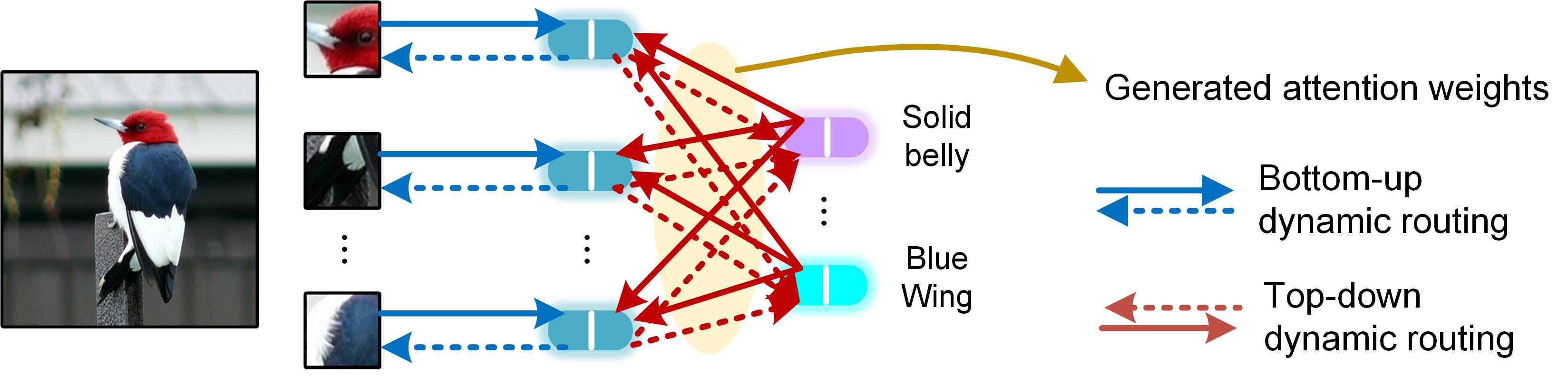}
	\caption{Brief illustration of the propose active attention mechanism, which contains both the bottom-up dynamic routing process and the top-down dynamic routing process.} \label{1}
\end{figure*}

Specifically, when trying to recognize an image from unseen classes, humans will involuntarily try to establish a connection between the attribute semantics with the corresponding local image regions. Besides, humans achieve semantic alignment by ruling out the irrelevant visual regions and locating the most relevant ones in a gradual way~\cite{ji2018stacked}. The above two phenomenons motivate us to introduce the top-down guidance and dynamic routing connection into the attention mechanism. As shown in Figure \ref{1} and bottom-left of Figure \ref{2}, the newly proposed active attention is constructed by both the bottom-up and the top-down connection pathways, and each pathway is formed by the dynamic routing rather than the conventionally used convolutional forward passing. Such an active attention works as the transformer encoder in our framework to learn the attribute-aligned visual features.

After obtaining the attribute-aligned visual features, we calculate the correlation between the attribute-aligned visual features and the corresponding attribute semantics, and generate the final class label predictions by multiplying the obtained correlation vector to the class attribute vectors (see bottom-right of Figure \ref{2}). When considering the attribute semantics as keys, the attribute-aligned visual features as queries, and the class attribute vectors as values, respectively, it is interesting to see that such a process can also be interpreted as a transformer decoding process that works in semantic space. As the input visual features of the transformer decoder have already been well aligned with the semantic space, the involved elements are connected with the static routing rather than the dynamic routing.

Based on the above-mentioned transformer encoder and decoder, we build a novel transformer model, called hybrid routing transformer (HRT). The overall learning model is shown in Figure \ref{2}. HRT mainly has three-fold novel properties: 1) From the perspective of the attention mechanism, it is not based on the commonly used passive self-attention. Instead, HRT is designed with the active semantic-guided attention. Particularly, the HRT encoder performs attention in both bottom-up and top-down manners. 2) From the perspective of the inner connection, HRT is formed by both the dynamic routing (in transformer encoder) and the static routing (in transformer decoder), which build different types of connection to the elements involved in the transformer. 3) From the perspective of zero-shot learning, we reveal an under-studied yet important issue, i.e., the active connection between the visual feature and attribute vector, and build the first transformer-based zero-shot image recognition framework.

\section{Related Work}

\textbf{Zero-Shot Learning.} Zero-shot learning (ZSL) aims to predict objects in unseen classes or both seen and unseen classes~\cite{xian2019f}, the former is called traditional ZSL while the later is called generalized ZSL (GZSL). The core is to transfer knowledge learned from seen classes to unseen classes. The existing methods can be divided into three types: (1) Embedding methods~\cite{akata2015label,akata2015evaluation,annadani2018preserving,liu2020hyperbolic,liu2018zero,xie2020region,xu2020attribute,xu2017matrix,GongMultimodal2016,zhang2019co,zhang2018zero,zhang2015zero}, which usually learn a compatibility function between image and class embedding spaces for similarity measurement. Specially, Akata \emph{et al.}~\cite{akata2013label} propose a bilinear-style hinge loss to learn the compatibility function. Based on~\cite{akata2013label}, Xian \emph{et al.}~\cite{xian2016latent} introduce non-linearity to ALE model. Following the structured SVM formulation, Akata \emph{et al.}~\cite{akata2015evaluation} designs a multiclass loss. Huynh \emph{et al.}~\cite{huynh2020fine} leverage attribute semantic vectors to learn the association between images and attributes. (2) Generative methods~\cite{yu2020episode,wu2020self,chen2018zero,felix2018multi,GongCentroid2020,kumar2018generalized,mikolov2013distributed}, which aim to generate synthetic samples of unseen classes from semantic information and then set the ZSL problem as a supervised classification problem. Common generative methods use GAN~\cite{goodfellow2014generative}, VAE~\cite{kingma2013auto} or flow-based generative models~\cite{dinh2014nice}. Xian \emph{et al.}~\cite{xian2018feature} directly generate image features conditioned on the class-level semantic descriptors. Felix \emph{et al.}~\cite{felix2018multi} generate synthetic features by a multi-modal cycle-consistent GAN. (3) Gating Methods~\cite{atzmon2019adaptive,chen2020boundary}, which use a gating based mechanism to separate the unseen samples from the seen samples for GZSL. Ideally, if the gate mechanism of binary classification is very effective, GZSL can be divided into a traditional ZSL problem to classify unseen samples and a supervised classification problem to classify seen samples. According to the experimental settings, the existing methods can be divided into two types: (1) Inductive ZSL, which only uses seen sample with labels in the training phase to classify unseen samples in the testing phase. (2) Transductive ZSL, which uses both seen samples with labels and unseen samples without labels in the training phase, which enables the model to use unseen visual features in the training phase to alleviate domain shift problem. Our experiments use the embedding method under the inductive ZSL setting.

\textbf{Attention in ZSl.} The aim of the attention mechanisms is to either highlight important local information or alleviate the influence of irrelevant and noisy information~\cite{ji2018stacked}. Ji \emph{et al.}~\cite{ji2018stacked} weight different local features by a stacked attention mechanism, with access to the costly part annotations during training. Zhu \emph{et al.}~\cite{zhu2019semantic} weight different global features by learning multiple channel-wise attentions. Xie \emph{et al.}~\cite{xie2019attentive} leverage attentive region embedding to learn the bilinear mapping to the semantic space. \cite{huynh2020fine} is the closest competitor, which uses the passive attention to build similarity between features and class attribute vectors. However, these works either need part annotations or use passive sematic-unguided attention mechanism, thus enormous sematic gap still exists between the two unrelated modalities of image and attribute.
In order to establish active correlation between image and attribute to capture discriminative features, we apply the capsule to transformer for ZSL, and build dynamic bottom-up and top-down attention mechanism by initializing high-level capsules with class-semantic vectors and performing low-level capsules with patch features. Our method proved to be very effective in subsequent experiments.


\textbf{Capsule-Transformer.} The transformer is proposed in~\cite{vaswani2017attention} with the attention-based encoder-decoder architecture. It is successfully used in the natural language processing field~\cite{bert2019transformer} firstly and then extended to computer vision tasks~\cite{dosovitskiy2020image,carion2020end}. Capsule network was first introduced by~\cite{detection2020transformer}, aiming to improve the ability of identifying spatial relationships and rotation of the CNN structure. Capsule is a group of neurons. Using the dynamic routing method between two capsule layers, capsule network can match CNN in recognition results. Although, there appears few works~\cite{chen2018Syntax, capsule2020transformer} to improve the transformer attention with capsule network for machine translation. The proposed HRT method has distinct properties with the existing models by using the active semantic-guided attention in both the bottom-up and top-down manner. It is also worth mention that this is the earliest work to establish a capsule-transformer-like framework for solving the ZSL problem.

\section{Hybrid Routing Transformer}

\subsection{Problem Setting and Overall Framework}

In zero-shot learning, we consider seen classes $\mathcal{C}_{s}$ and unseen classes $\mathcal{C}_{u}$,  where $\mathcal{C}_{s} \cap \mathcal{C}_{u}=\varnothing$. Specifically, we denote the training data as $\mathcal{D}^{s}=\left\{\left(x_{i}, y_{i},\mathbf z_{i} \right)\right\}$, where $x_{i}$ and $y_{i}$ denote the training image and the corresponding label, and $\mathbf z_{i}=[z_1, ..., z_A]$ represents the associated class attribute vectors. The target data contains images with both seen classes and unseen classes as well as the semantic vectors for each class.

\begin{figure*}[t]
	\centering
	\includegraphics[width=1.0\textwidth]{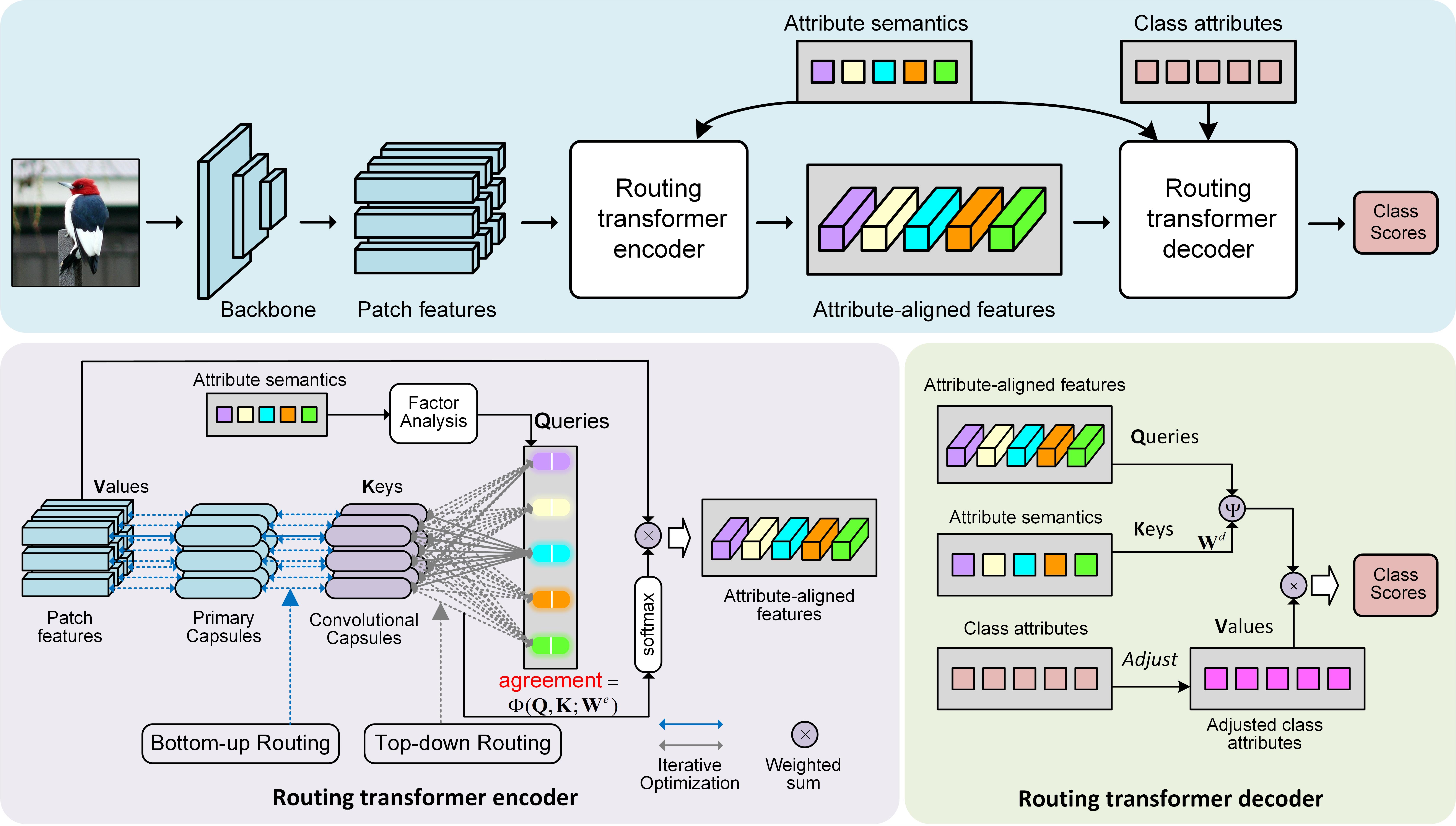}
	\caption{Our Hybrid routing transformer consists of an \textsl{Semantic-guided dynamic transformer routing Part} as the encoder constructing the crucial semantic-aligned visual features of an image and an \textsl{Semantic-guided static transformer routing Part} as the decoder transferring the attribute-aligned features into classification scores under the guidance of class attribute vectors. Both of the two parts are trained jointly to establish the necessary top-down guidance and semantic alignment.} \label{2}
\end{figure*}

In this work, given an input image, we first extract the basic feature representation for each image patch by ResNet-101. Then, we solve ZSL by a newly-designed transformer, where the transformer encoder extracts attribute-aligned visual features, while the transformer decoder projects the attribute-aligned visual features into classification scores. In particular, by designing bottom-up and top-down, dynamic and static routing layers in the transformer architecture, we build hybrid routing transformer to perform active interaction between images and attributes.

\subsection{HRT Encoder}
Given the patch features $\{\mathbf f^r\}_{r=1}^R$ of each input image, we equip HRT Encoder with attribute-guided dynamic routing layers to obtain the attribute-aligned visual features. Fistly, in order to obtain a compact description of each image patch, we cluster every patch feature into one capsule of a smaller dimension $d=16$ by a bottom-up routing process, creating new patch capsule $\mathbf g^r \in \mathbb R^{1 \times d}$. In specific, given one patch feature vector $\mathbf f^r \in \mathbb R^{1\times 2048}$, we convert it to primary capsules, each of which is represented by a $4\times4$ pose matrix, through a $1\times1$ learnable convolution on $\mathbf f^r$ and obtain 128 primary capsules for each image region as illustrated in Figure \ref{2}. Then, we use a bottom-up routing process to facilitate the routing between the 128 primary (child) capsules and one parent capsule. The bottom-up routing is implemented based on the EM routing~\cite{hinton2018matrix}. For child capsule $i$, firstly it is transformed by $\mathbf T_{i j}$ to cast a vote $\mathbf O_{i j} = \mathbf M_{i} \mathbf T_{i j}$ for the parent capsule $j$. Then the non-linear routing process is performed by the EM algorithm, where the vectorized version of $j^{th}$ pose matrix $\mathbf M_j$ is the expectation of $j^{th}$ Gaussian distribution. Denote $P_{i \mid j}$ as the probability density of the vectorized vote $\mathbf O_{i j}$ under $j^{th}$ Gaussian model. $P_{i \mid j}^{h}$ is $h^{th}$ component of $P_{i \mid j}$, which has variance $\left(\sigma_{j}^{h}\right)^{2}$ and mean $\mu_{j}^{h}$. The $h^{th}$ component of the probability density $P_{i \mid j}^{h}$ is computed as,
\begin{equation}
P_{i \mid j}^{h}=\frac{1}{\sqrt{2 \pi\left(\sigma_{j}^{h}\right)^{2}}} \exp \left(-\frac{\left(\mathbf O_{i j}^{h}-\mu_{j}^{h}\right)^{2}}{2\left(\sigma_{j}^{h}\right)^{2}}\right).
\end{equation}

Then we calculate the activation of the parent capsule $j$ based on the minimum description length principle :
\begin{equation}
\begin{split}
\alpha_{j}=& \operatorname{logistic}\left(\lambda\left(\beta-\gamma \sum_{i} r_{i j}-\sum_{h} \operatorname{cost}_{j}^{h}\right)\right), \\
&\cos t_{j}^{h}=-\sum_{i} r_{i j} \ln P_{i \mid j}^{h},
\end{split}
\end{equation}

where $\beta$ and $\gamma$ are two learnable parameters, $\cos t_{j}^{h}$ indicates the cost for activating the parent capsule $j$. $\sum_{i} r_{i j}$ calculates the amount of child capsules assigned to the parent capsule $j$. $\lambda$ is a hyper-parameter. In EM routing, M-step and E-step run iteratively. The M-step computes the outputs of pose matrix and activation of the parent capsule $j$, and the E-step exports the possibility of child capsules assigned to the parent capsule $j$. We obtain the parent capsule $\mathbf{g}^r$ as the compact representation for each image patch by iteratively calculating the pose matrix $\mathbf{M}$ between the child and parent capsules in the EM routing process.

In order to attain active guidance of attribute semantics, we then use a top-down routing process to establish connections between the obtained image patch capsules and the global attribute semantic capsules transformed from the original attribute semantic vectors. Similar to~\cite{huynh2020fine}, we apply GloVe~\cite{pennington2014glove} to extract the $\tau$-dimensional attribute semantic vectors $\{{\mathbf v}_a\}_{a=1}^A$, which is followed by a demission reduction process based on the Factor Analysis~\cite{loehlin1987latent}. Finally, we obtain the $d$-dimensional compact attribute semantic vectors $\{\widetilde{\mathbf{v}}_a\}_{a=1}^A$, where $a \in [1,2, \cdots, A]$ is the attribute index. To build the top-down routing connection, we adopt the Inverted Dot-Product Attention routing process~\cite{tsai2020capsules}. Such routing process first calculates the vote $\nu_{i j}=\mathbf{W}_{i j}^{e} \cdot \mathbf{p}_{i}$ for the child capsule $\mathbf{p}_{i}$. Then it computes the agreement as $\mathbf{o}_{i j}=\mathbf{p}_{j}^{\top} \cdot \nu_{i j}$ by the dot-product similarity between parent capsule $\mathbf{p}_{j}$ and the vote. Then we update the parent capsules $\mathbf{p}_{j}$ by:
\begin{equation}
\mathbf{p}_{j}=\text { LayerNorm }\left(\sum_{i} r_{i j} \nu_{i j}\right),  r_{i j}=\frac{\exp \left(\mathbf{o}_{i j}\right)}{\sum_{j^{\prime}} \exp \left(\mathbf{o}_{i j^{\prime}}\right)}.
\end{equation}
The above routing process will be performed several times to strength the agreement between the child and parent capsules.

It worth mentioning that, the parent capsules $\mathbf{p}_{j}$ in our framework are initialized with $\{\tilde{\mathbf v}_a\}_{a=1}^A$, while other capsule networks perform zero or random initialization on them. The advantage of our approach is that it can realize an attribute-guided and active attention mechanism between the two-modality information. In fact, such a routing process simultaneously facilitates the semantic feature embedding and the semantic coherence computing. Under this circumstance, we treat the attribute semantic vector as attribute query $\mathbf Q$, the patch capsule as visual key $\mathbf K$. Then, the agreement between the child and parent capsules can properly reflect the similarity or the relationship between $\mathbf Q$ and $\mathbf K$. Since the agreement is implicitly parameterized by $\mathbf W^{e}$, we denote it as $\Phi(\mathbf Q, \mathbf K;\mathbf W^{e})\in \mathbb R^{R \times A}$. Then, the final attribute-aligned visual features $\mathbf H=[\mathbf h_1, \mathbf h_1, \cdots, \mathbf h_A]\in \mathbb R^{2048 \times A}$ can be obtained by:
\begin{equation}
\mathbf H=HRT^E(\mathbf Q,\mathbf K,\mathbf V) = \operatorname{softmax}({\Phi(\mathbf Q, \mathbf K;\mathbf W^{e})})\mathbf V\label{eq:1},
\end{equation}
where $\mathbf V\in \mathbb R^{2048 \times R}$ indicates the value matrix, which is formed by the original $R$ patch features $\{\mathbf f^r\}$.

\subsection{HRT Decoder}

Given the attribute-aligned visual features $\mathbf H$, we equip HRT Decoder with semantic-guided static routing layers to obtain the final class probability.

In the fine-grained recognition, there are so many attribute scores for classes while only a small portion of the attributes are crucial to distinguish different classes. In order to focus on the important attributes, we adjust the $c$-th class attribute vectors belonging by $\tilde{\mathbf z}^c = \operatorname{sigmoid}({\Lambda^{T} \mathbf{W}_{\beta} \mathbf {H}\odot\mathbf {L}}) \mathbf z^c$, where $\Lambda=[\mathbf v_1, \mathbf v_2, \cdots, \mathbf v_A]\in \mathbb R^{\tau \times A}$, $\mathbf {L}$ is the unitary matrix, $\odot$ indicates the element-wise production. As the decoding process works in semantic space, we regard the attribute-aligned visual features in $\mathbf {H}$ as the queries $\mathbf Q\in \mathbb R^{2048 \times A}$, the attribute semantic vectors $\{\mathbf v_a\}_{a=1}^A$ as the keys $\mathbf K\in \mathbb R^{\tau \times A}$, while the adjusted class attribute vectors $\{\tilde{\mathbf z}^c\}_{c=1}^C$ as the values $\mathbf V\in \mathbb R^{A \times C}$. Considering that the attribute-aligned visual features and the attribute semantic vectors have a clear correspondence relationship, we define $\Psi (\mathbf Q, \mathbf K;\mathbf W^{d})=\mathbf I_{1\times A}\operatorname{diag}(\mathbf Q^T \mathbf{W}^{d} \mathbf K)$ to represent the content-aware attribute vectors, where $\mathbf{W}^{d}\in \mathbb R^{2048 \times \tau}$ is an embedding matrix between $\mathbf Q$ and $\mathbf K$. Then, the final class scores of the input image can be obtained by measuring the coherence between the content-aware attribute vectors and the adjusted class attribute vectors:
\begin{equation}
\mathbf s=HRT^D(\mathbf Q,\mathbf K,\mathbf V) = \Psi (\mathbf Q, \mathbf K;\mathbf W^{d})\mathbf V\label{eq:2}.
\end{equation}

\subsection{Training and Testing}
For training the proposed deep model, we leverage three-fold loss function:
\begin{equation}
\mathcal{L}_{HRT}=\mathcal{L}_{ce}+\lambda_{1} \mathcal{L}_{cal}+\lambda_{2} \mathcal{L}_{reg},\label{eq:10}
\end{equation}
where $\lambda_{1}$ and $\lambda_{2}$ are hyper-parameters. The first loss $\mathcal{L}_{ce}$ is the cross-entropy loss, which measures the consistency between the classification prediction and the ground-truth label. The second one is the calibration loss~\cite{huynh2020fine}  $\mathcal{L}_{cal}= -\sum_c y^c \log p(s^{c}+\gamma_{c})$, where $p\left(s^{c}\right) = \frac{\exp \left(s^{c}\right)}{\sum_{c^{\prime} \in \mathcal{C}} \exp \left(s^{c^{\prime}}\right)}$, $s^{c}$ is the prediction of the c-th class and $y^c$ is its corresponding label, $\gamma_{c}$ is the hyper-parameter to balance the prediction score between the seen and unseen classes, which is different for seen and unseen classes. This loss aims to alleviate the biased prediction towards the seen categories~\cite{huynh2020fine}. The third one is the attribute regression loss $\mathcal{L}_{reg}=\|{\varphi(\mathbf Q, \mathbf K;\mathbf W^{d})}-\mathbf z^{c*}\|_{2}^{2}$, where $c*$ indicates the ground-truth class index. This loss constraints the predict attribute score vector to agree with the class attribute vector that is obtained by human statistic, which can help to improve the generalization capacity to the unseen classis.

In testing, we use $c^{*}=\operatorname{argmax}_{c \in \mathcal{C}} (s^{c}+\gamma_{c})$ to obtain the predicted class belonging to either the seen classes or unseen ones.

\section{Experiments}

\subsection{Setting}\label{Setting}

\textbf{Datasets.} We evaluate the performance of our model on three widely used ZSL benchmark datasets: \textsl{Caltech-UCSD Birds-200-2011} (CUB)~\cite{welinder2010caltech}, \textsl{SUN attributes} (SUN)~\cite{patterson2014sun} and \textsl{Animals with Attributes 2} (AWA2)~\cite{xian2017zero}. The CUB ~\cite{welinder2010caltech} is a fine-grained dataset containing 11788 bird images from 150 seen classes and 50 unseen classes with 312 expert-annotated attributes. Although it contains discriminative attribute location annotation to distinguish fine-grained classes, our model works under the weakly supervised setting without the part annotations unlike~\cite{ji2018stacked}. SUN ~\cite{patterson2014sun} is another fine-grained dataset containing 14204 scene images, from 645 seen classes and 72 unseen classes with 102 annotated attributes. Different from the above two datasets, AWA2 ~\cite{xian2017zero} dataset is in the coarse-grained level consisting of 37322 animal images, from 40 seen classes and 10 unseen classes with 85 attributes.

\textbf{Evaluation Metrics.} In this work, we measure the average per-class top-1 accuracy of our method on both traditional ZSL and GZSL setting. For traditional ZSL task, we evaluate test images only from unseen classes with T1 (top-1 accuracy).  For GZSL task, we evaluate test images from both seen classes and unseen classes. Following the protocol proposed in~\cite{xian2017zero}, we report \textsl{tr} and \textsl{ts} as the average per-class top-1 accuracy of test images from seen classes and unseen classes, respectively. \textsl{H} is computed as the hamonic mean of \textsl{tr} and \textsl{ts} to measure the comprehensive performance, which can be calculated by $H \triangleq 2 \times \frac{tr \times ts}{tr+ts}$.

\textbf{Visual features.} We obtain the patch features at the last convolution layer of ResNet-101~\cite{he2016deep} model pre-trained on the ImageNet-1K~\cite{deng2009imagenet} dataset. For CUB, it contains so many similar sub-categories under a big category of birds which needs much abundant visual information to classify them. Therefore, in order to obtain richer visual features, we modify the stride of $\operatorname{conv}5_{-} x$ layer in ResNet-101 from 2 to 1, and get the patch features of size $14\times14\times2048$. The differences among the categories in SUN and AWA2 are relatively large, finer features may confuse the correspondence between visual features and attributes, and eventually deteriorate classification accuracy. Thus, we only adjusted the stride of feature extraction process for CUB dataset.

\textbf{Implementation Details.} Following ~\cite{xian2017zero}, we adopt ResNet-101~\cite{he2016deep} pre-trained on ImageNet-1K~\cite{deng2009imagenet} as the backbone for feature extraction without fine-tuning. Given the input image of size  $224\times 224$, we will obtain patch features of size $7\times7\times2048$ or $14\times14\times2048$ for different datasets. We use RMSprop~\cite{tieleman2012lecture} optimization method by setting momentum as 0.9, weight decay of $10^{-4}$ and the initial learning rate of $10^{-3}$. The coefficients $\lambda_{1}$ and $\lambda_{2}$ in Eq.(\ref{eq:10}) are set as 0.1 and 0.033, respectively. The factor $\gamma_{c}$ in calibration loss for seen classes is different, which is -0.5 for CUB and SUN dataset, -0.8 for AWA2 dataset, while $\gamma_{c} = 1$ for unseen classes on all the three datasets. The model is implemented based on PyTorch platform~\cite{paszke2019pytorch}, training on a single 2080 Ti GPU card.

\subsection{Comparison with the state-of-the-art}\label{Comparison}

In the experiment, we compare our proposed HRT method with several state-of-the-art embedding methods on both ZSL and GZSL settings. We report the top-1 accuracy and harmonic mean of each method in Table \ref{table1}, where ``-" indicates that the results are not reported. These methods, CONSE~\cite{norouzi2013zero}, DEVISE~\cite{frome2013devise}, ALE~\cite{akata2013label}, SJE~\cite{akata2015evaluation}, ESZSL~\cite{romera2015embarrassingly}, SSE~\cite{zhang2015zero}, SYNC~\cite{changpinyo2016synthesized}, LATEM~\cite{xian2016latent}, SAE~\cite{kodirov2017semantic}, learn a compatibility function between visual image and attributes for similarity measurement. Meanwhile, these methods, SGMA~\cite{zhu2019semantic}, LFGAA~\cite{liu2019attribute}, AREN~\cite{xie2019attentive} and DAZLE~\cite{huynh2020fine}, introduce the attention mechanism to embedding methods. D-VAE~\cite{li2021generalized} and GCM-CF~\cite{yue2021counterfactual} are generative models. Extensive experiments demonstrate the effectiveness of our proposed method, and the experiment results on all the three datasets show that our proposed method, achieve superior performances to the existing state-of-the-art methods in most cases.

\footnotetext[1]{SEM of HRT means the Standard Error of mean accuracy of our model under 5 random seed experiments.}
\begin{table*}\centering
\caption{Comparison of our method with the state-of-the-art embedding methods on CUB, SUN and AWA2.  The methods with * denote fine-tuning the backbone weights, otherwise fixing them. We measure top-1 accuracy(T1) in ZSL setting, and top-1 accuracy on seen/unseen (tr/ts) classes and their harmonic mean (H) in GZSL setting. Bold font denote the best results.}

\label{table1}
\resizebox{\textwidth}{!}{
\begin{tabular}{@{}cc|ccc|ccc|ccc|ccc@{}}
\toprule
  & \multirow{3}{*}{Method} & \multicolumn{3}{c|}{Zero-shot Learning} & \multicolumn{9}{c}{Generalized Zero-shot Learning}                           \\
  \cline{3-14}
  &                                   & AWA2        & CUB         & SUN        & \multicolumn{3}{c}{AWA2} & \multicolumn{3}{c}{CUB} & \multicolumn{3}{c}{SUN} \\ \cline{3-14}
  &                                   & T1          & T1          & T1         & tr     & ts     & H      & tr     & ts     & H     & tr     & ts     & H     \\ \hline
  & CONSE~\cite{norouzi2013zero}      & 44.5        & 34.3        & 38.8       & 90.6   & 0.5    & 1.0    & 72.2   & 1.6    & 3.1   & 39.9   & 6.8    & 11.6  \\
  & DEVISE~\cite{frome2013devise}     & 59.7        & 52          & 56.5       & 74.7 & 17.1  & 27.8   & 53.0   & 23.8   & 32.8  & 27.4   & 16.9   & 20.9  \\
  & ALE~\cite{akata2013label}         & 62.5        & 54.9        & 58.1       & 81.8   & 14     & 23.9   & 62.8   & 23.7   & 34.4  & 33.1   & 21.8   & 26.3  \\
  & SJE~\cite{akata2015evaluation}    & 61.9        & 53.9        & 53.7       & 73.9   & 8.0    & 14.4   & 59.2   & 23.5   & 33.6  & 30.5   & 14.7   & 19.8  \\
  & ESZSL~\cite{romera2015embarrassingly}& 58.6        & 53.9        & 54.5       & 77.8   & 5.9    & 11.0   & 63.8   & 12.6   & 21    & 27.9   & 11.0   & 15.8  \\
  & SSE~\cite{zhang2015zero}        & 61.0        & 43.9        & 51.5       & 82.5   & 8.1    & 14.8   & 46.9   & 8.5    & 14.4  & 36.4   & 2.1    & 4.0   \\
  & SYNC~\cite{changpinyo2016synthesized}& 46.6        & 55.6        & 56.3       & 90.5   & 10.0   & 18.0   & 70.9   & 11.5   & 19.8  & \textbf{43.3}   & 7.9    & 13.4  \\
  & LATEM~\cite{xian2016latent}       & 55.8        & 49.3        & 55.3       & 77.3   & 11.5   & 20.0   & 57.3   & 15.2   & 24    & 28.8   & 14.7   & 19.5  \\
  & SAE~\cite{kodirov2017semantic}    & 54.1        & 33.3        & 40.3       & 82.2   & 1.1    & 2.2    & 54.0   & 7.8    & 13.6  & 18.0   & 8.8    & 11.8  \\
  & D-VAE~\cite{li2021generalized}    & -           & -           & -          & 80.2   & 56.9   & 66.6   & 58.2   & 51.1   & 54.4  & 47.6   & 36.6   & 41.4 \\
  & GCM-CF~\cite{yue2021counterfactual}   & -           & -           & -         & 75.1   & \textbf{60.4}   & 67.0   & 59.7   & 61.0   & 60.3  & 37.8   & 47.9     & \textbf{42.2}  \\
  & SGMA*~\cite{zhu2019semantic}      & \textbf{68.8}        & 71.0        & -          & 87.1   & 37.6   & 52.5   & 71.3   & 36.7   & 48.5  & -      & -      & -     \\
  & LFGAA*~\cite{liu2019attribute}& 68.1        & 67.6        & 61.5       & \textbf{93.4}   & 27.0   & 41.9   & \textbf{80.9}   & 36.2   & 50.0  & 40.4   & 18.5   & 25.3  \\
  & AREN*~\cite{xie2019attentive}& 67.9        & \textbf{71.8}        & 60.6       & 92.9   & 15.6   & 26.7   & 78.7   & 38.9   & 52.1  & 38.8   & 19.0   & 25.5  \\
  & DAZLE~\cite{huynh2020fine}        & 67.9        & 65.9        & 59.3       & 75.7   & 60.3   & 67.1   & 59.6   & 56.7   & 58.1  & 24.3   & 52.3   & 33.2  \\ \cline{2-14}
  & HRT(ours)           & 67.3        & 71.7        & \textbf{63.9}       & 78.7   & 58.9   & \textbf{67.4}   & 63.5   & \textbf{62.1}   & \textbf{62.8}  & 26.9   & \textbf{53.2}   & 35.7 \\
  & SEM of HRT \footnotemark[1]     & $\pm$0 & $\pm$0.01 &$\pm$0.04 & $\pm$0.01 & $\pm$0.02 & $\pm$0.02 & $\pm$0.01 & $\pm$0.08 & $\pm$0.03 & $\pm$0.05 & $\pm$0.01 & $\pm$0.04 \\ \bottomrule

\end{tabular}
}
\end{table*}

\textbf{Generalized Zero-shot Learning Results.} As shown in Table \ref{table1}, the proposed HRT method acheives superior results on AWA2, CUB and SUN datasets for generalized ZSL task in most cases. We gain impressive results for the harmonic mean (H), where 62.8\% on CUB, and 67.4\% on AWA2. On SUN dataset, we obtains 53.2\% for top-1 accuracy on unseen classes, which is also much better than other methods. In particular, our method significantly outperforms other algorithms on unseen accuracy, surpassing the state-of-the-art model on unseen accuracy by 4.4\% on CUB dataset. However, since our capsule attention model is a kind of the dense attention model, lacking the training samples makes the proposed method cannot achieve the best harmonic mean on SUN dataset, where this dataset contains only 16 training samples for each seen classes. Furthermore, DAZLE~\cite{huynh2020fine} is the closest competitor, which uses the passive attention to build similarity. It causes confused relationship and cannot bridge the semantic gap between the visual image and the attribute descriptions. Different from it, we construct an active semantic-guided attention mechanism which helps high-level semantic actively focus on relevant visual features. As a result, we achieve superior results, improving harmonic mean (H) of 4.7\% on CUB and 2.5\% on SUN dataset absolutely, compared to DAZLE~\cite{huynh2020fine}.
Notice that LFGAA*~\cite{liu2019attribute} achieves the best seen accuracy, while our HRT model obtains the best unseen accuracy with relatively high seen accuracy.
Moreover, the experiment results in Table~\ref{table1} show that our proposed HRT model can balance the performance of both seen and unseen classes greatly, compared with these methods, AREN*~\cite{xie2019attentive}, LFGAA*~\cite{liu2019attribute}, SGMA*~\cite{zhu2019semantic}, LATEM~\cite{xian2016latent}. This maybe due to that our proposed HRT method constructs an active attention mechanism and facilitates a more effective knowledge transfer from seen classes to unseen classes.

\textbf{Zero-shot Learning Results.} In ZSL, training classes are disjoint with testing classes. As shown in Table \ref{table1}, our proposed method presents a significant improvement compared with the state-of-the-art methods in most cases, i.e., 2.4\% on SUN. The SUN dataset is a scene dataset, our method is able to understand the abstract content of scenes more deeply, through constructing the hybrid dynamic top-down and bottom-up attention pathways between visual image and the attribute. On CUB and AWA2 datasets, we gain comparable performance with the best method AREN*~\cite{xie2019attentive} and SGMA*~\cite{zhu2019semantic}.

\subsection{Ablations}

\begin{table*}\centering
\caption{Performance analysis of our HRT framework with CNN and the capsule network in the transformer encoder on CUB, SUN and AWA2, we denote them as HRT-CNNNet and HRT-CapsuleNet respectively. The methods HRT-CNNNet and HRT-CapsuleNet are trained only with cross-entropy and calibration loss, while HRT-Ours is trained by the whole loss in Eq~\ref{eq:10}. We measure top-1 accuracy(T1) in ZSL setting, and top-1 accuracy on seen/unseen (tr/ts) classes and their harmonic mean (H) in GZSL setting. }

\label{table3}
\resizebox{\textwidth}{!}{
\begin{tabular}{@{}cc|ccc|ccc|ccc|ccc@{}}
\toprule
  & \multirow{3}{*}{Method} & \multicolumn{3}{c|}{Zero-shot Learning} & \multicolumn{9}{c}{Generalized Zero-shot Learning}                           \\
\cline{3-14}
  &                                   & AWA2        & CUB         & SUN        & \multicolumn{3}{c}{AWA2} & \multicolumn{3}{c}{CUB} & \multicolumn{3}{c}{SUN} \\ \cline{3-14}
  &                                   & T1          & T1          & T1         & tr     & ts     & H      & tr     & ts     & H     & tr     & ts     & H     \\ \hline

  & HRT-CNNNet            & 66.9        &64.3        &59.0       &75.9  & 59.3   &66.6   &59.4   &54.8   &57.0  &23.6   &51.8   &32.4 \\
  & HRT-CapsuleNet        & 67.7       & 69.3        & 60.4       & 75.3   & 60.2   & 66.9   & 62.4   &60.0   &61.2  & 25.8   & 50.8   & 34.2  \\
  & HRT-Ours            & 67.7        & 71.7        & 63.9       & 78.7   & 58.9   & 67.4   & 63.5   & 62.1   & 62.8  & 26.9   & 53.2   & 35.7 \\ \bottomrule
\end{tabular}
}
\end{table*}

To illustrate the effectiveness of the proposed capsule network based HRT encoder framework, we make detailed performance comparison between the traditional CNN based HRT encoder and the capsule network based encoder framework on CUB, SUN and AWA2 datasets as shown in Table~\ref{table3}. We denote them as HRT-CNNNet and HRT-CapsuleNet, respectively. The methods HRT-CNNNet and HRT-CapsuleNet are trained only with cross-entropy and calibration loss, while HRT-Ours is trained by the whole loss in Eq~\ref{eq:10}. We can clearly see that the capsule network based encoder framework is better than that of the traditional CNN based HRT encoder framework, HRT-CapsuleNet outperforms the baseline HRT-CNNNet by 2.1\% under the H(harmonic mean) measurement in the GZSL setting, while 2.4\% under the T1 measurement in the ZSL setting.

\begin{table*}[t]\centering\small
\caption{Ablation results for (generalized) zero-shot learning on CUB dataset under the same baselines.}

\label{table2}
\begin{tabular}{@{}cccc|cccc@{}}
\toprule
\begin{tabular}[c]{@{}c@{}} Hybrid Routing \\ Transformer Encoder\end{tabular} &\begin{tabular}[c]{@{}c@{}} Hybrid Routing \\ Transformer Decoder\end{tabular} & $\mathcal{L}_{cal}$ &$\mathcal{L}_{reg}$ & tr     & ts     & H     & T1     \\ \hline
\ding{55}  & \ding{55}   & \ding{55}  & \ding{55}                      & 59.7   & 1.9    & 3.7   & 51.4   \\
\ding{51}  & \ding{55}   & \ding{55}  & \ding{55}                      & 66.4   & 9.6    & 16.8  & 63.1   \\
\ding{51}  & \ding{51}   & \ding{55}  & \ding{55}                      & 69.6   & 10.0   & 17.5  & 67.7   \\
\ding{51}  & \ding{51}   & \ding{51}  & \ding{55}                      & 62.4   &60.0   &61.2  & 69.8   \\
\ding{51}  & \ding{51}   & \ding{51}  & \ding{51}                      & 63.5   & 62.1   & 62.8  & 71.7   \\ \bottomrule
\end{tabular}
\end{table*}

To further measure the influences of different components in HRT, we perform ablation study on CUB dataset under both ZSL and GZSL settings. As shown in Table \ref{table2}, the first line means that we train HRT only with cross-entropy loss and the patch features in hybrid routing transformer encoder and decoder are performed without any attention mechanism. By adding hybrid routing transformer encoder, hybrid routing transformer decoder, $\mathcal{L}_{cal}$ loss and $\mathcal{L}_{reg}$ loss gradually, the GZSL results demonstrate that these four components improve accuracy by a huge margin over baseline, i.e., 3.8\% (tr), 60.2\% (ts), 59.1\% (H) and 20.3\% (T1), absolutely. The second line means that we perform the semantic-guided hybrid routing transformer in encoder, and the harmonic mean is significantly improved by 13.1\%, which is achieved by constructing both the bottom-up and the top-down dynamic routing pathways to generate the aligned features. The third line represents that by adding the semantic-guided static transformer routing in decoder, and the model further improves the performances. In addition, the calibration loss is an essential part to boost accuracy under GZSL setting. Finally, through attribute regression loss, we further strengthen supervision and correct the classification results, boosting the harmonic mean by 3.4\%. The above four components constitute the model together, improving the performance under both ZSL and GZSL settings.

\subsection{Hyper-Parameter Selection}

\begin{figure*}[t]
    \centering
    \begin{minipage}[t]{0.5\linewidth}
    \centering
    \includegraphics[width=3in]{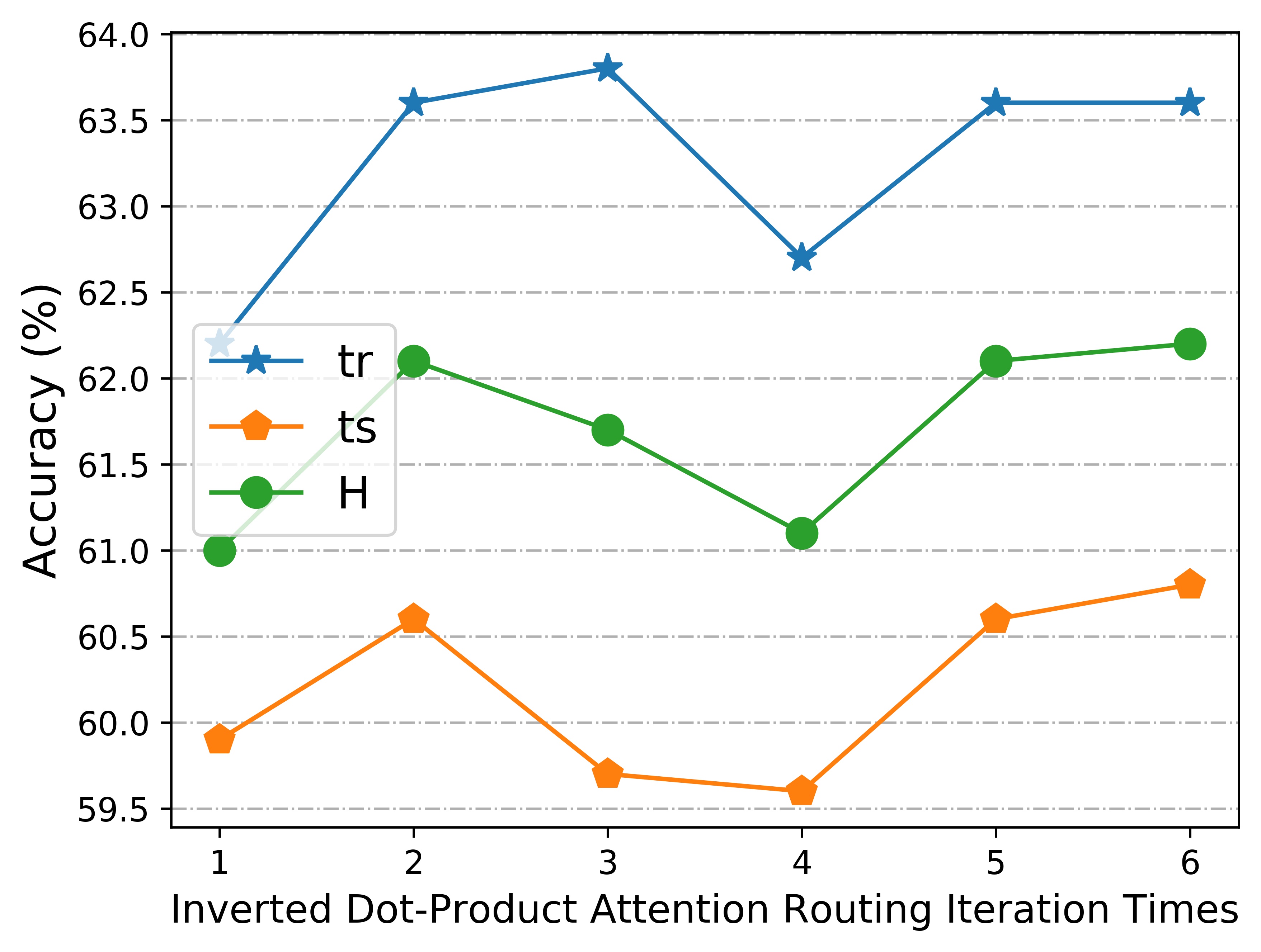}
    \end{minipage}%
    \begin{minipage}[t]{0.5\linewidth}
    \centering
    \includegraphics[width=3in]{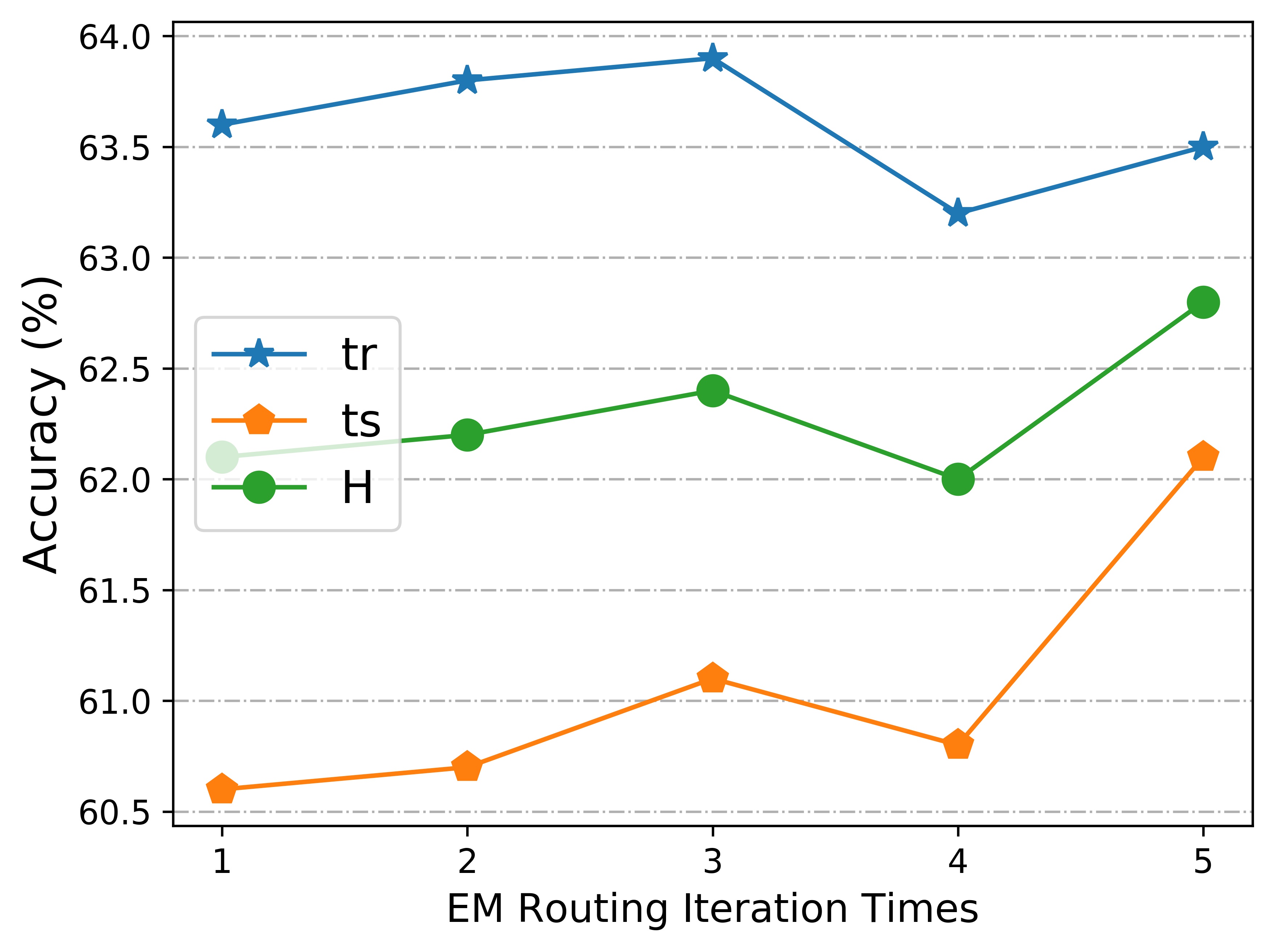}
    \end{minipage}%
\centering
\caption{(a) shows the changes of tr, ts and H with different Inverted Dot-Product Attention routing iteration times under CUB dataset while EM routing iteration time is equal to 1. (b) displays the result with variational EM routing iteration times while Inverted Dot-Product Attention routing iteration time is set to 2.}\label{3}
\end{figure*}

For the purpose of establishing the necessary top-down guidance and semantic alignment for attending to the real attribute-correlation regions, we build the HRT model via bottom-up and top-down routings. To observe the influence of the iteration times of these two routings, we conduct experiments with various iteration values, i.e.$\{1,2,3,4,5\}$. General ZSL results on CUB dataset are illustrated in Figure~\ref{3}. In Figure~\ref{3} (a), we set the EM routing iteration time to 1, and the top-down dynamic attention routing iteration time changes from 1 to 5. We get the best results with the top-down dynamic attention routing iteration of 2. In Figure~\ref{3} (b), we fix the top-down dynamic attention routing iteration time to 2 and study the performances under different EM routing iteration times. We obtain the best results with EM routing iteration of 5. The above experiment results demonstrate the superiority of the proposed dynamic bottom-up and top-down routing mechanism, which helps HRT model establish an active connection between the visual feature and attribute vector rather than a simple passive mapping.

\subsection{Qualitative results}

\begin{figure*}
	\centering
	\includegraphics[width=1\textwidth]{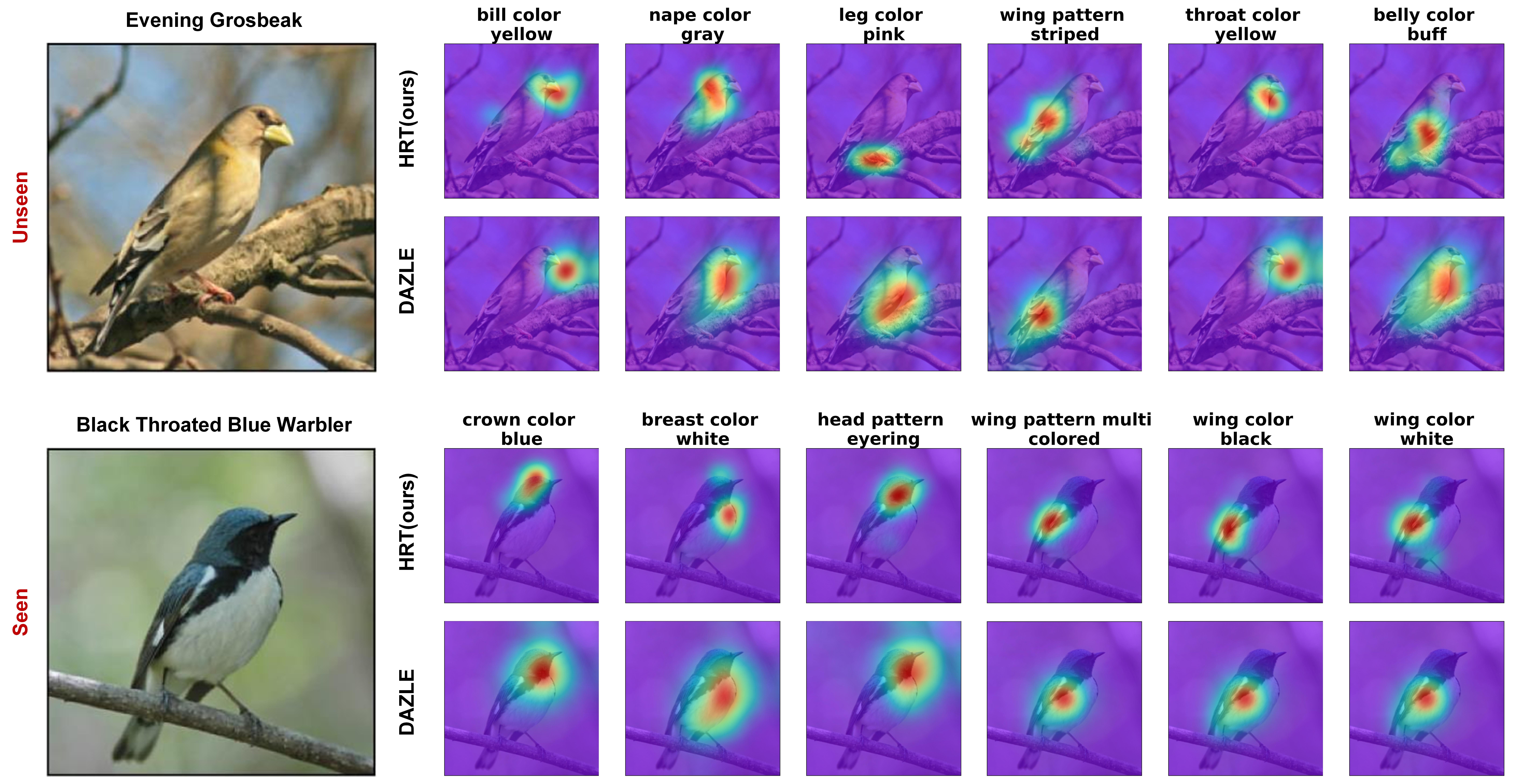}
	\caption{Visualization of attribute-aligned feature locating for examples from unseen and seen classes on CUB dataset.} \label{4}
\end{figure*}

We compared the feature visualization results of the proposed HRT model with the baseline passive attention model DAZLE\cite{huynh2020fine}, by contacting image feature and attribute on the CUB dataset. Figure \ref{4} visualizes the agreement maps for examples from unseen and seen classes. On the whole, the attention maps generated by the HRT method are more concentrated and contacting image feature and attribute more accurately. For the bird ``Evening Grosbeak'' of unseen classes, DAZLE tends to generate inaccurate or deflected attention maps covering the whole bird. For example, with the DAZLE model, the attributes ``Bill Color Yellow'', ``Throat Color Yellow'', ``Belly Color Buff'' and ``Leg Color Pink'' concentrate on the false visual information, and the attribute ``Wing Pattern Striped'' dosen't capture relevant visual information. For the bird ``Black Throated Blue Warbler'' of the seen classes, though DAZLE can generate attention maps utilizing the learned knowledge, it can't capture the true attribute-aligned features and its attention maps are more dispersive. On the other hand, the attention maps of our HRT model focuses on the attribute-aligned visual features more accurately. These examples demonstrate that our hybrid routing transformer model constructs an active and effective connection between image feature and attribute, and thus alleviate the semantic gap effectively between two different modality information for zero-shot learning.

\section{Conclusions}

In this work, we propose a hybrid routing transformer (HRT) framework for ZSL and GZSL tasks. We are the first to apply the routing-based transformer for ZSL, by constructing semantic-guided active mechanism to alleviates the semantic gap between the visual modality and attribute modality effectively, and facilitate knowledge transfer. The proposed HRT model is a novel encoder-decoder framework. In the HRT encoder part, we utilize both the bottom-up and top-down dynamic routing pathways to generate attribute-aligned visual features. While in the HRT decoder part, we take the semantic-guided static routing to transfer attribute-aligned features into classification scores under the guidance of class attribute vectors. Extensive experiments suggests that the proposed active bottom-up and top-down dynamic routing pathway can help improve the transformer, thus we can extend our method to other research field in the future.

\bibliographystyle{IEEEtran}
\bibliography{IEEEexample}


\end{document}